\def\year{2020}\relax
\documentclass[letterpaper]{article} 
\usepackage{aaai20}  
\usepackage{times}  
\usepackage{helvet} 
\usepackage{courier}  
\usepackage[hyphens]{url}  
\usepackage{graphicx} 
\usepackage{graphicx}  
\usepackage{paralist}
\usepackage{amsmath}
\usepackage{bm}
\usepackage{amssymb}

\usepackage{subcaption}
\usepackage{wrapfig}
\usepackage{hhline}
\usepackage{enumitem}
\usepackage{booktabs}
\usepackage[title]{appendix}

\usepackage{color,soul}
\usepackage{algorithm}
\usepackage{algorithmic}
\usepackage{graphicx}
\usepackage{multirow}
\usepackage{array}
\newcolumntype{C}[1]{>{\centering\let\newline\\\arraybackslash\hspace{0pt}}m{#1}}
\newcommand{\citet}[1]{\citeauthor{#1} \shortcite{#1}}
\newtheorem{lemma}{Lemma}
\makeatletter
\newcommand{\printfnsymbol}[1]{%
  \textsuperscript{\@fnsymbol{#1}}%
}
\makeatother

\title{Light Multi-segment Activation for Model Compression}

\author{Zhenhui Xu,\textsuperscript{\rm 1}\thanks{Equal Contribution. This work was done while the first author was visiting Microsoft Research.} 
Guolin Ke,\textsuperscript{\rm2}\printfnsymbol{1} 
Jia Zhang,\textsuperscript{\rm 2} 
Jiang Bian,\textsuperscript{\rm 2} 
Tie-Yan Liu\textsuperscript{\rm 2}\\
\textsuperscript{\rm 1}Peking University, 
\textsuperscript{\rm 2}Microsoft Research \\
zhenhui.xu@pku.edu.cn,
\{guolin.ke, jia.zhang, jiang.bian, tie-yan.liu\}@microsoft.com}

\begin{document}

\maketitle

\begin{abstract}
Model compression has become necessary when applying neural networks (NN) into many real application tasks that can accept slightly-reduced model accuracy but with strict tolerance to model complexity. Recently, Knowledge Distillation, which distills the knowledge from well-trained and highly complex teacher model into a compact student model, has been widely used for model compression. However, under the strict requirement on the resource cost, it is quite challenging to make student model achieve comparable performance with the teacher one, essentially due to the drastically-reduced expressiveness ability of the compact student model. Inspired by the nature of the expressiveness ability in NN, we propose to use multi-segment activation, which can significantly improve the expressiveness ability with very little cost, in the compact student model. Specifically, we propose a highly efficient multi-segment activation, called Light Multi-segment Activation (LMA), which can rapidly produce multiple linear regions with very few parameters by leveraging the statistical information. With using LMA, the compact student model is capable of achieving much better performance effectively and efficiently, than the ReLU-equipped one with same model complexity. Furthermore, the proposed method is compatible with other model compression techniques, such as quantization, which means they can be used jointly for better compression performance. Experiments on state-of-the-art NN architectures over the real-world tasks demonstrate the effectiveness and extensibility of the LMA.

\end{abstract}

\section{Introduction}

Neural Network (NN) has become a widely-used model in many real-world tasks, such as image classification, translation, speech recognition, etc. In the meantime, the increasing size and complexity of the advanced NN models have raised a critical challenge~\cite{Wang2018Private} in applying them into many real application tasks, which can accept appropriate performance drop with very extremely-limited tolerance to high model complexity. Running NN models on mobile devices and embedded systems are emerging examples that make every effort to avoid expensive computation and storage cost but can endure slightly-reduced model\footnote{Unless otherwise stated, the term ``model'' used in this paper refers to the Neural Network model.} accuracy.  

Consequently, many studies have been paying attention to producing compact and fast NN models with maintaining acceptable model performance. 
In detail, there are two active directions investigated model compression through pruning \cite{lecun1990optimal,hassibi1993second,han2015learning,li2016pruning,frankle2018the} or quantizing \cite{courbariaux2015binaryconnect,rastegari2016xnor,mellempudi2017ternary} the trained large NN models into squeezed ones with trimmed redundancy but preserved accuracy. More recently, increasing efforts explored Knowledge Distillation \cite{hinton2015distilling} to obtain compact NN models by training them with the supervision from well-trained larger NN models~\cite{Polino2018Model,Wang2018Private,mishra2018apprentice,hubara2017quantized,luo2016face,wu2016quantized,zhu2016trained,sau2016deep}. 
Compared with directly training a compressed model from scratch merely using the ground truth, the supervision in terms of soft distributed representations on the output layer of the large teacher model can even significantly enhance the effectiveness of the resulting compact student model.
In practice, nevertheless, it is quite difficult to produce the compressed student model that can yield similar effectiveness to the complex teacher model, essential due to the limited expressiveness ability of the compressed one in terms of the strictly-restricted parameter size.

\begin{figure}
    \centering
    \includegraphics[width=0.9\linewidth]{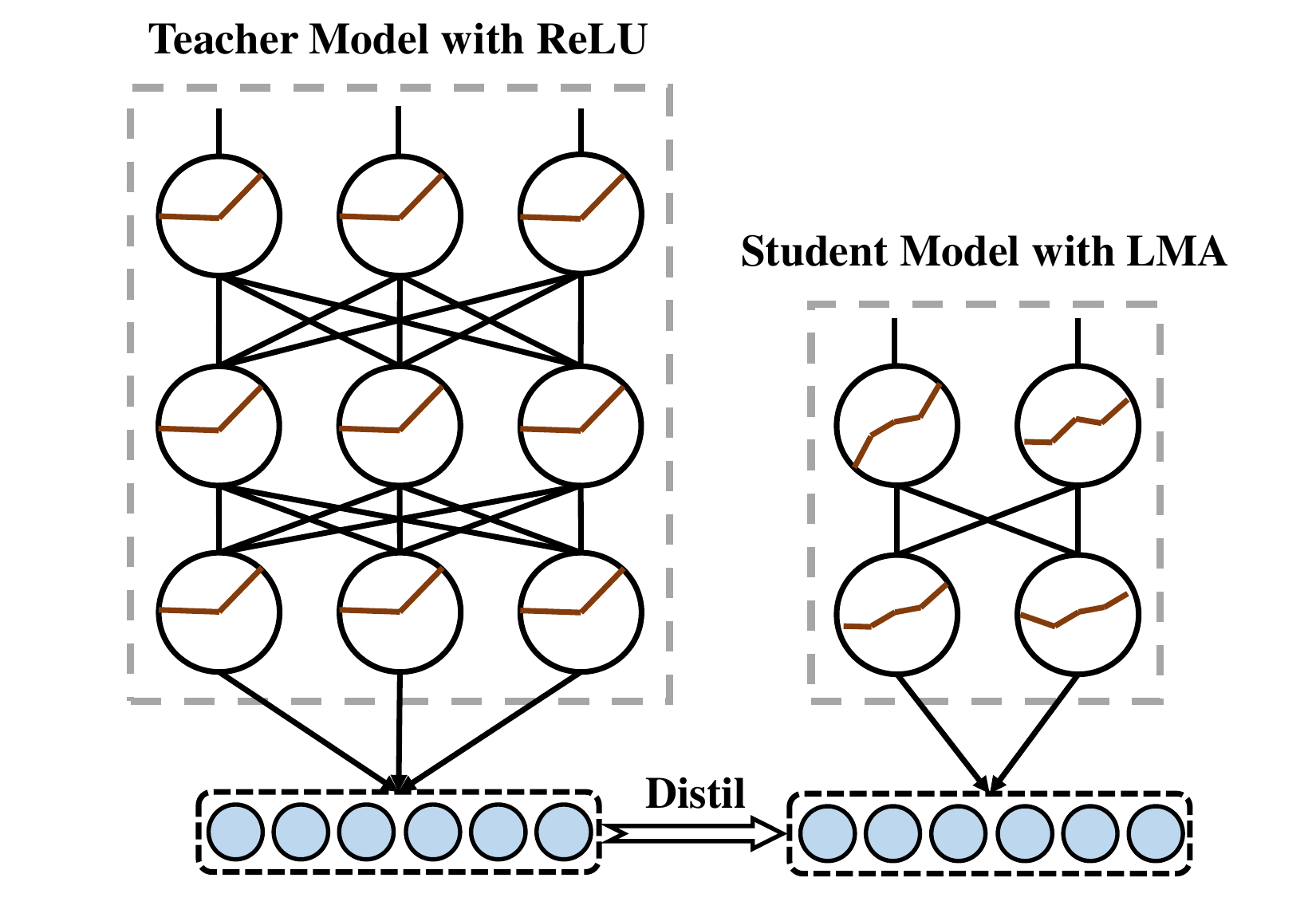}
    \caption{Depiction of knowledge distillation with LMA.}
    \label{fig:MC}
\end{figure}

Intuitively, to enhance the power of the compressed model, it is necessary to increase its expressiveness ability. However, traditional approaches to introduce more layers or hidden units into the model can easily violate the strict restrictions on the model size.
Fortunately, besides the neuron number, the nonlinear transformation, in terms of the activation, within the NN model plays an equally-important role in reflecting the expressiveness ability. As pointed out by \cite{montufar2014number}, an NN model that uses multi-layer ReLU or other piecewise linear activation as the activation is still essentially a complex piecewise linear function. Moreover, the number of linear regions produced by an NN model depends on not only its model size but also its activations. The more segments in piecewise linear activation, the more regions NN will produce, but there are only two segments in widely-used ReLU. Inspired by the few costs in adding more regions in the activation, it is more reasonable to improve the expressiveness ability of NN via multi-segment activations, instead of increasing the model size, either depth or width.

Thus, in this paper, we introduce a novel highly efficient piecewise linear activation, in order to improve the expressiveness ability of the compressed models with little cost. 
In detail, as shown in Fig.~\ref{fig:MC}, we leverage a generic knowledge distillation framework for model compression, in which, however, the compact student model is equipped with proposed multi-segment piecewise linear activations, named Light Multi-segment Activation (LMA). 
In LMA, we first cut the input range into multiple segments based on batch statistics information, and ensure it can adapt to any range of input lightly and efficiently. Then, we assign each input with customized slope and bias depending on the segment it belongs to, which thus empower NN models higher expressiveness ability due to the stronger non-linearity of the new activation.
Owing to the above design, LMA-equipped compact student models yield two advantages: 1) It has much higher expressiveness ability, compared with one merely endowed with vanilla ReLU; 2) Its resource cost is still much smaller and even more controllable compared with the other type of multi-segment piecewise linear activation.

Extensive experiments of multi-size NN architectures on various real tasks, including image classification and machine translation, have demonstrated both the effectiveness and the efficiency of LMA, which implies the improved expressiveness ability, thus the performance, of the LMA-equipped model.
Additional experiments further illustrate that our method can also improve the expressiveness ability of the models that have been compressed by some other popular techniques, such as quantization, so jointly using the others and ours can achieve even better compression results.

The main contributions of this paper are multi-fold:
\begin{itemize}
    \item To the best of our knowledge, it is the first work that leverages multi-segment piecewise linear function as activation in model compression. It proposes a novel multi-segment activation, which improves the expressiveness ability of the compressed student model within the knowledge distillation framework.
    \item With using statistical information of each batch, the proposed activation can efficiently improve the performance of compressed models with preserving low resource cost.
    \item The proposed method is compatible with the other popular compression techniques, such that it is easy to combine them and further get better compression results.
    \item On various real challenging tasks, experimental results of multiple models with different sizes show our methods have good performance. And the effectiveness of joint usage, that combines our method with the others, is also shown in the experiments.
\end{itemize}

\section{Related work}

This work is mainly related to two research areas, model compression and piecewise linear activation. The representative work of the former includes model pruning, quantization and distillation, while the latter typically studies respective effects of ReLU, Maxout and APLU on the performance of NN.

\subsection{Model Compression}
In this area, \cite{lecun1990optimal,hassibi1993second} first explored pruning based on second derivations. More recently, \cite{han2015learning,han2016dsd,jin2016training,hu2016network,yang2017designing} pruned the weights of Neural Networks with different strategies and made some progress. Most recently, \cite{frankle2018the} showed a dense Neural Network contains a sparse trainable subnetwork such that it can match the performance of the original network, named as the lottery ticket hypothesis. On the other hand, \cite{gupta2015deep} have done a comprehensive study on the effect of low precision fixed point computation for deep learning.
Therefore, quantization is also an active research area, where various methods were proposed by many works \cite{mellempudi2017ternary,hubara2017quantized,mellempudi2017ternary,zhu2016trained,rastegari2016xnor,wu2016quantized}.

Besides, using distillation for size reduction is mentioned by \cite{hinton2015distilling}, which gives a new direction for training compact student models. The weighted average of soft distributed representation from the teacher's output and ground truth is much useful when training a model, so some practices \cite{Wang2018Private,luo2016face,sau2016deep} have been put for training compressed compact model. Moreover, recent works also proposed to combine the quantization with distillation, producing better compression results. Among these, \cite{mishra2018apprentice} used knowledge distillation for low-precision models, which proposes distillation can also help training the quantized model. \cite{Polino2018Model} proposed a more in-depth combination of these two methods, named Quantized Distillation.
Besides, there are also some works \cite{han2015deep,iandola2016squeezenet,wen2016learning,gysel2016hardware,mishra2017wrpn} further reduced the model size by combining multiple compression techniques like quantization, weight sharing and weight coding. Similarly, the combination of our method with the other is also shown in this paper.

\subsection{Piecewise Linear Activation}
A piecewise linear function is composed of multiple linear segments. Some piecewise functions are continuous when the boundary value calculated by two adjacent intervals function is the same, whereas some may not be continuous. Benefit from its simplicity and the fitting ability to any function with enough segments,
it is widely-used in machine learning models \cite{landwehr2005logistic,malash2010piecewise}, especially as activations in Neural Networks \cite{lecun2015deep}. 
Theoretically, \cite{montufar2014number,pascanu2013number} studied the number of linear regions in Neural Networks produced by piecewise linear activation functions (PLA), which can be used to measure the expressiveness ability of the networks. 

Specifically, as a two-segment PLA, Rectified Linear Unit (ReLU) \cite{nair2010rectified} and its parametric variants can be generally defined as $h_i(x) = \min(0, a _i x ) + \max(0, x) ~$, where $x$ is the input, $a _i$ is a linear slope, and $h_i(x)$ is the activated output. For original ReLU, it fixes $a_i$ to zero so the formula degenerates to $h_i(x) = \max(0, x)$; Parametric ReLU (PReLU) \cite{he2015delving} makes $a_i$ learnable and initializes it to 0.25.
Besides, there are also some PLAs with multiple segments improved from ReLU. For example, Maxout \cite{goodfellow2013maxout} is a typical multi-segment PLA, which is defined as $h_i(x)=\max(z_{ij})$ for all $j\in[1,k]$, where $k$ can be treated as its segment number, and it transforms the input into the maximum of $k$-fold linear transformed candidates $z_{ij}$; Adaptive Piecewise Linear Units (APLU) \cite{agostinelli2014learning} is also a multi-segment one, which is defined as a sum of hinge-shaped functions,
\begin{equation}\label{eq:aplu} 
\small
    h_i(x) = \max(0,x)+\sum\nolimits_{j=1}^k a_i^j \max(0, -x+b_i^j) ~, 
\end{equation}
where $k$ is a hyper-parameter set in advance, while the variables $a_i^j$, $b_i^j$ for $j \in \{1, ..., k\}$ are learnable. The $a_i^j$ control the slope of the linear segments while the $b_i^j$ determine the locations of the hinges similar to segments. 

In this paper, after studying the connection of above two areas, we are the first to leverage the properties of PLA for model compression, that to improve the expressiveness ability of compact model via multi-segment activation, thereby improving its performance.

\section{Methodology}

We start by studying the connection between PLA and the expressiveness ability of Neural Networks, followed by introducing the Light Multi-segment Activation (LMA) that is used to further improve the performance of the compact model in model compression. 

\subsection{Preliminaries}

\subsubsection{Expressiveness Ability Study}
Practically, increasing complexity of the neural networks, in terms of either width \cite{zagoruyko2016wide} or depth \cite{he2016deep}, can result in swelling performance, essentially due to the higher expressiveness ability of the NN.
However, when applying the NN into some resource-exhausted environments, its size cannot be inflated without limit.
Fortunately, the nonlinear transformation within the NN, in terms of the activation, provides another vital channel to enhance the expressiveness ability.
Yet, the widely-used ReLU in NN is just a simple PLA with only two segments, where the slope on the positive segment is fixed to one while the other is zero.
Therefore, other than enlarging the size of the NN model, another effective alternative method to enhance the expressiveness ability of the NN model is to leverage more powerful activation functions.
In this paper, we propose to increase the segment number in activation to enhance its expressiveness ability, and further empower the compact NN to yield good performance.

Theoretically, there are also some related analysis \cite{montufar2014number} that can justify our motivation. As pointed out by them, the capacity, i.e. the expressiveness ability, of a PLA-activated Neural Network can be measured by the number of linear regions of this model.
And for an NN, in the $l$-th hidden layer with $n_l$ units, the number of separate input-space neighbourhoods that are mapped to a common neighborhood $R \subseteq S_l \subseteq \mathbb{R}^{n_l}$ can be decided recursively as
\begin{equation}
\label{NRL}
\small
    \mathcal{N}_R^l = \sum\nolimits_{R' \in P_R^l} \mathcal{N}_{R'}^{l-1},~\mathcal{N}_R^0 = 1,~~~\text{for each}~R \subseteq \mathbb{R}^{n_0},
\end{equation}
where $S_{l}$ denotes the set of (vector valued) activations reachable by the $l$-th layer for all possible input; $P_R^l$ denotes the set of subsets $\bar{R}_1,...,\bar{R}_k \subseteq S_{l-1}$ that are mapped by the activation onto $R$. Based on the above result, the following lemma (see \cite{montufar2014number}; Lemma 2) is given.
\begin{lemma}
The maximal number of linear regions of the functions computed by an $L$-layer Neural Network with piecewise linear activations is at least $\mathcal{N} = \sum_{R \in P^L}\mathcal{N}_{R}^{L-1}$~, where $\mathcal{N}_{R}^{L-1}$ is defined by Eqn.~(\ref{NRL}), and $P^L$ is a set of neighborhoods in distinct linear regions of the function computed by the last hidden layer.
\end{lemma}

Given the above lemma, the number of linear regions of a Neural Network is in effect influenced by the layer number, the hidden unit size, and the region number in PLA. From ReLU to Maxout, the significant improvement is on the $P^l$ in the lemma, which is also the nature of our approach.
Taking Maxout as an example of detailed analysis, it can lead to an important corollary that a Maxout network with $L$ layers of width $n$ and rank $k$ can compute functions with at least $k^{L-1}k^{n}$ linear regions (see \citet{montufar2014number}; Theorem 8). Meanwhile, ReLU can be treated as a special rank-2 case of Maxout, whose bound is obtained similarly by \cite{pascanu2013number}. Obviously, the number of linear regions can be improved by increasing either $L$, $n$ or $k$. 
However, in a compressed model, neither the layers $L$ nor hidden units $n$ can be increased too much. Thus, we propose to construct a highly efficient multi-segment activation function with its linear regions $k$ becomes larger.


\subsubsection{Analysis on Existing Multi-segment PLAs}
As mentioned in Related Work, some previous studies have already proposed some multi-segment PLAs. In the following of this subsection, we will analyze whether they are suitable for being applied in model compression. 

Considering Maxout first, its regions are produced by $k$-fold weights and only the maximum of its $k$-fold outputs is picked to feed forward, which obviously causes the redundancy within Maxout.
On the contrary, to construct a PLA with multiple segments and ensure limited parameters increment in the meantime, a more intuitive inspiration, from the definition of piecewise linear function, lies in that it first cuts the input range into multiple segments, and then transforms the input linearly by individual coefficients (i.e. slopes and biases) on different segments.
In this way, the parameter number of the network based on this scheme can be controlled as $ L*(k+n^2)$~, compared with $L*kn^2$ in the above assumed Maxout NN.

In fact, APLU is a hinge-based implementation of this scheme, with few additive parameters. Specifically, in Eqn.~\ref{eq:aplu}, $\bm{b}$ are the cut points of the input range, and $\bm{a}$ can be grouped accumulatively into coefficients. However, APLU can increase the memory cost due to its accumulation operation. In details, APLU requires $k$ times intermediate variables to compute the items parallel and then accumulates all of them one-time. Although we also accumulate them recursively to avoid this, it will be $k$ times slower and is unacceptable. Besides, with the $k$ becomes larger, the memory cost will growth linearly. 

In a word, neither Maxout nor APLU can be directly employed for model compression in that Maxout produces much more parameters and APLU is memory-consuming. In the following subsection, we will introduce a new activation process that is both effective and efficient for model compression.

\subsection{Light Multi-segment Activation}

\subsubsection{Method}
LMA mainly contains two steps. The first is batch segmentation, which is proposed to find the segment cut-points based on the batch statistical information. Then the inputs are transformed with the corresponding linear slopes and biases according to their belonging segments.

Firstly, to construct a multi-segment piecewise activation, it needs to cut the continuous inputs to multiple segments. 
There are two straightforward solutions: 1) pre-defined like the vanilla ReLU; 2) training cut-points like APLU. For the former, as the input ranges of hidden layers are dramatically changed during training, it is hard to define the appropriate cut points in advance. For the latter, the cut points are unstable due to the random initialization and stochastic update by back-propagation. As the naive solutions cannot work well, inspired by the success of Batch Normalization \cite{ioffe2015batch}, we propose Batch Segmentation, to determine the segment boundaries by the statistical information. 

There are two statistical schemes \cite{dougherty1995supervised} to find appropriate segments. One is based on frequency, and the other one is based on numerical values.
Concretely, after using the frequency-based method, each segment has the same number of inputs, while if using numerical value-based one, the numerical width of each segment is equal.
Indeed, the frequency-based method is more robust since it is not sensitive to numerical values. However, it is not efficient, especially running on GPU and applied for model compression. Thus, the numerical value-based solution is used in LMA for efficiency purpose. 
Specifically, we assume the input is a normal distribution and cut the segments by equal value width. So, here each segment cut-point is defined as,
\begin{equation} \label{eq:value_bs}
\small
    \begin{aligned}
    b_0 = \mu - 3\sigma,~~~~~~~~
    b_j = b_{j-1} + \frac{ 6\sigma}{k},~\text{for}~ j=1, 2,\ldots,k ~,
    \end{aligned}
\end{equation}
where $k$ is the segment number, a hyper-parameter, $\mu$ and $\sigma$ are the mean and standard deviation of the batch input $\bm{x}$, respectively. To reduce the effect of outliers and make use of the property of normal distribution, we assume $\mu \pm 3\sigma$ are the range endpoints and assign cut points according to this assumption. 
Like Batch Normalization, the moving average of $\bm{b}$ is used in the test phase. To further improve the efficiency, as well as more stable statistical information, the $\bm{b}$ could be calculated and shared in the same layer.


After determining segment boundaries, it needs to assign the coefficient, i.e. slope and bias, to each input according to its belonging segment. 
To avoid the memory-consuming problem in APLU, we use the independent slopes and biases in LMA.
Formally, the activation process can be defined as,
\begin{equation}
\label{eq:LMA}
\small
h_i(x) = \alpha^i_j \cdot x + \beta^i_j, ~~~ x \in (b_{j},b_{j+1}]
\end{equation}
where $\bm{\alpha}$ denotes the slope coefficient, $\bm{\beta}$ denotes the bias, and $j$ denotes segment indices. Especially, considering there still may be few extreme inputs out of the normal distribution assumption, the first and last segment are set to $(-\infty,b_1]$ and $(b_{k-1},+\infty)$ respectively, instead of determining by $b_0$ and $b_k$. Finally, after the above steps, the linear transformed values $h_i(x)$ feed-forward to the next layer.

\begin{table}
\caption{Cost comparison between multi-segment activation functions.}
\label{wrap-tab:cc}
\centering
\begin{tabular}{cccc}\\\toprule
~ & Maxout & APLU  & LMA \\ \midrule
Param. Size & $O(k*n^2)$ & $O(k*n+n^2)$ & $O(k+n^2)$\\  
Mem. Cost & $O(k*n)$ & $O(k*n)$ & $O(n)$\\  \bottomrule
\end{tabular}
\end{table} 

\subsubsection{Analysis and Discussion}
In the following, we will take more detailed discussions on LMA from the perspective of complexity analysis and initialization.
Obviously, in LMA, there is only two additional trainable variables $\bm{\alpha}$ and $\bm{\beta}$ for each layer, whose total size is $2 * k * n$, where $k$ is segment number and $n$ is hidden unit number. Furthermore, to reduce the parameter size extremely, the $\bm{\alpha}$ and $\bm{\beta}$ are shared in the layer-level, which means that all the units or feature maps are activated by the same LMA in one specific layer. Therefore, the parameters brought by LMA in one layer is only $2*k$, even reduced by $n$ times compared with APLU. Moreover, about the running memory cost in inference phase, LMA only produces the belonging segment indices for inputs, whose space cost is $O(n)$, while APLU needs $O(k*n)$ hinges and Maxout needs $O(k*n)$ activation candidates. To conclude, the cost comparisons between each multi-segment PLAs are shown in Table~\ref{wrap-tab:cc}, where the parameter size and the running space cost at activation in one layer are listed. It shows LMA is more suitable for model compression because of its less storage and running space cost.


Besides, the slopes and biases on all segments need to be initialized in LMA. The initialization methods always can be categorized into two classes: 1) random initialization like the other parameters in NN; 2) initializing it as a known activation, such as vanilla Relu or PReLU. Though the random initialization does not impose any assumptions and may achieve a better performance \cite{mishkin2015all}, it usually introduces uncertainty and leads to unstable training too. With this in mind, we choose the second initialization method for LMA. Specifically, we initialize the LMA to be the vanilla ReLU, which means that all biases are initialized to zero, the slopes of the half left segments are initialized to zero while the rest slopes are initialized to one.


\subsubsection{Model Compression}
As an effective method to improve the expressiveness ability of the compressed model, LMA can be applied with distillation and other compression techniques.
Under the distillation framework, we first train a state-of-the-art model and get as much good performance. Then given it as the teacher model, a more compact architecture is employed to as the student to learn the knowledge from the teacher. Because of the parameter reduction in the student, it always underperforms much lower than the teacher despite using knowledge distillation. Here, we replace all the original ReLUs with our LMA for the student model, improving its expressiveness ability, and further improving the performance much. The replacement is very convenient that it only needs to change one line of code in the implementation. After that, according to \cite{hinton2015distilling,Polino2018Model}, the distillation loss for training the student is also a normal weighted average of the loss from ground truth and the one from teacher's output, which is formally defined as,
\begin{equation}
\small
    \mathcal{L} = (1-\alpha) \mathcal{L}_{CE/NLL}(\hat{y}(\bm{x}), ~y_{GT}) + \alpha \mathcal{L}_{KL}(\hat{y}(\bm{x}), ~y_{T}) ~,
\end{equation}
where $\alpha$ is a hyper-parametric factor, which is always set to 0.7, to adjust the weight of two losses; $\hat{y}(\bm{x})$ is the student's output logits; the first loss $\mathcal{L}_{CE/NLL}$ is a Cross Entropy Loss or Negative Log Likelihood Loss with the ground truth labels $y_{GT}$, depending on the tasks (CE is for image classification and NNL is for machine translation in our experiments); the latter loss $\mathcal{L}_{KL}$ is a Kullback–Leibler Divergence Loss with the teacher's output logits $y_{T}$. Additionally, when calculating $\mathcal{L}_{KL}$, we also use a temperature factor $\tau$ to soften the $y_T$ and the $\hat{y}$, whose specific settings will be shown in the experiments.

Besides, LMA is well compatible with the other compression techniques, since it is convenient to replace the activations from ReLU with LMA. For example, based on a recent representative method, Quantized Distillation \cite{Polino2018Model}, after replacing the ReLU with LMA in student model, though it is quantified to low-precision model during training, our method still empowers it to achieve higher performance than origin one, which will be shown in the experiments.

\begin{table*}
    \centering
    \small 
    \caption{Image Classification Results. 
    The metrics of Teacher models on each dataset are shown in the left-most cells. The accuracy (\%) is shown in ``\textit{mean $\pm$ std}~'' pattern and the inference memory cost (in MB) is shown in ``\textit{A (+D)}'' pattern, where \textit{A} denotes absolute memory cost and \textit{D} is additional part compared with ReLU-equipped model. The last column shows the improvement of LMA (comparing performance with ReLU and memory with APLU). \textbf{The results show the PLA with more segments (APLU and LMA) outperforms the fewer ones, especially widely-used ReLU, meanwhile LMA maintains much lower memory cost than APLU.} \textsuperscript{*}For a \textbf{Maxout-8}-based Student Model, whose size is the same as Student 1, only get 87.76$\pm$0.55\% accuracy, even lower than ReLU-based one.}
    \begin{tabular}{c|c|c|c c c|c c c}
    \toprule
     \multicolumn{3}{c|}{\textbf{Method}} & \textbf{ReLU} & \textbf{PReLU} & \textbf{Swish} & \textbf{APLU-8} & \multicolumn{2}{c}{\textbf{LMA-8}}\\
     \hline
     \multirow{4}*{CIFAR-10} & Student 1 & Acc. 
     & 88.74 $\pm$0.25 
     & 89.31 $\pm$0.35 
     & 89.03 $\pm$0.11 
     & 89.92 $\pm$0.21 
     & \textbf{90.57 $\pm$0.20} & 2.1\%~$\uparrow$\\
     \cline{3-9}
     \multirow{4}*{21.4 MB} & 4.04 MB & Mem. & 14.24 & 15.95 (+1.7) & 15.10 (+0.9) & 25.80 (+11.6) & 16.81 (+2.6) & 78\%~~$\downarrow$\\
     \hhline{~|========}
     \multirow{4}*{Acc. 92.83} & Student 2 & Acc. 
     & 82.67 $\pm$0.46 
     & 84.35 $\pm$0.37 
     & 84.06 $\pm$0.36 
     & 85.31 $\pm$0.60 
     & \textbf{85.66 $\pm$0.34} & 3.6\%~$\uparrow$\\
     \cline{3-9}
     \multirow{4}*{Mem. 29.28} & 1.28 MB & Mem. & 3.40 & 4.72 (+1.3)  & 4.06 (+0.7)  & 11.69 (+8.3)  & 5.57 (+2.2) & 73\%~~$\downarrow$\\
     \hhline{~|========}
     ~ & Student 3 & Acc. & 73.33 $\pm$0.79 & 75.30 $\pm$0.17 & 75.45 $\pm$0.34 & 77.54 $\pm$0.97 & \textbf{77.66 $\pm$0.47} & 5.9\%~$\uparrow$\\
     \cline{3-9}
     ~ & 0.44 MB & Mem. & 1.45  & 2.07 (+0.6)  & 1.76 (+0.3)  & 5.15 (+3.7)  & 2.55 (+1.1) & 70\%~~$\downarrow$\\
     \hhline{=========}
     \multirow{1}*{CIFAR-100} & Student 1 & Acc. & 69.11 $\pm$0.80 & 70.03 $\pm$0.21 & 69.67 $\pm$0.40 & \textbf{70.99 $\pm$0.42} & 70.92 $\pm$0.42 & 2.6\%~$\uparrow$\\
     \cline{3-9}
     
     \multirow{1}*{68.7 MB} & 4.88 MB & Mem. & 16.27  & 16.40 (+0.13)  & 16.33 (+0.06)  & 17.03 (+0.76) & 16.46 (+0.19)  & 75\%~~$\downarrow$\\
     \hhline{~|========}
     
     \multirow{1}*{Acc. 77.56} & Student 2 & Acc. & 63.12 $\pm$1.00 & 64.52 $\pm$0.67 & 63.82 $\pm$0.78 & 66.28 $\pm$0.49 & \textbf{66.31 $\pm$0.68} & 5.1\%~$\uparrow$\\
     \cline{3-9}
     \multirow{1}*{Mem. 140.2} & 1.28 MB & Mem. &6.37  &6.44 (+0.07) & 6.41 (+0.04)  & 6.91 (+0.54) & 6.47 (+0.10) & 81\%~~$\downarrow$\\
    \bottomrule
    \end{tabular}
    \label{tab:IC-res}
\end{table*}

\section{Experiment}

In this section, we will conduct thorough evaluations on the effectiveness of LMA for model compression under two popular scenarios, image classification and machine translation. Besides, under the model compression scenarios, we will compare the performance of LMA with that of several widely-used baseline activations.\footnote{We released the code at: \url{https://github.com/motefly/LMA}} The baselines adopted includes ReLU\cite{nair2010rectified}, PReLU \cite{he2016deep}, APLU \cite{agostinelli2014learning} and Swish \cite{ramachandran2017swish}. For Swish is a well-known SOTA ReLU-like activation, we also adopt it to further show our effectiveness. Specifically, we will start with our experimental setup, including the data and models employed in the experiments. After that, we will analyze the performance of our method applied singly or jointly with some others to demonstrate its effectiveness and advantages for model compression.

\subsubsection{General Settings}
To ensure credible results, we run all the experiments 5 times with different random seeds, and report the average and standard deviation of them.
Besides, for fair comparisons, we set all the common parameters, including learning rate, batch size, hyper-parameters in distillation loss, etc., the same for all the baselines.
Note that, the settings for parametric baseline activations (PReLU, APLU and Swish), are all consistent with the original authors' demonstration.
For multi-segment activations (APLU and LMA), the segment numbers are set as the same to each other, which is 8 in our main experiments.
Moreover, to measure the resource cost by the models, we report their parameter size and inference memory cost (\textit{Mem.}), in which the latter is recorded when predicting the testing samples one by one. 
The model size hardly changes after replacing the activation function, since the additional parameters in all these activations are relatively very few. However, for another activation, Maxout, it will yield much more parameters if replaced. Due to the poor performance compared with other baselines under the setting of the same model size, we only report one little result (see Table~\ref{tab:IC-res}). 
More details about various specific parameter setting, model specifications and convergence curves can be found in the reproducibility supplementary materials.
 
\subsection{Image Classification}

\subsubsection{Settings}
Following what \cite{Polino2018Model} does in its code\footnote{\url{https://github.com/antspy/quantized_distillation}}, we first evaluate our method on CIFAR-10 and CIFAR-100, both of which are well-known image classification datasets. For experiments on CIFAR-10, some relatively small CNN architectures are employed, including one teacher model and three student models with different sizes.
Widen Residual Networks (WRN) \cite{zagoruyko2016wide} are employed for experiments on CIFAR-100, where WRN-16 is used as teacher model while two WRN-10 are used as students.
In the first phase, we train the teacher models and save them for the next distilled training. Then, we compare the performance of the student models with different activations under the supervision from both the teacher models and the ground truth. Accuracy (\textit{Acc.}) is used as the evaluation metric on this task.

\subsubsection{Result}
Table~\ref{tab:IC-res} summarizes the image classification results by various methods. From this table, we can find that the multi-segment activations (APLU and LMA) outperform the other baselines, on both two datasets with all the models of various sizes, where LMA outperforms ReLU by 2\% to 6\% on accuracy. Meanwhile, we can find that smaller compact model can imply more obvious improvement caused by multi-segment activations. In detail, on CIFAR-10, the LMA outperforms ReLU by 2\% on Student 1 while that is 6\% on Student 3. Besides, comparing APLU with LMA, we can easily find though their accuracy is sometimes close, the additional inference memory cost brought by equipping APLU is much larger than that by LMA, about 3 to 4 times more.

\begin{table*}[t]
    \centering
    \small 
    \caption{Machine Translation Results (\textit{Mem.} in MB).
    The metrics of Teacher models are shown in the left-most cells.
    Note that on WMT13, the memory needed for training APLU-equipped Student-1 exceeds the maximum memory of our GPU (24GB), thus there is no result of APLU. \textbf{Besides some similar observations on images, it also shows APLU on translations may also cost so much memory that the task failed, but LMA still works well.}}
    \begin{tabular}{c|c|c|c c c|c c c}
    \toprule
 \multicolumn{3}{c|}{\textbf{Method}} & \textbf{ReLU} & \textbf{PReLU} & \textbf{Swish} & \textbf{APLU-8} & \multicolumn{2}{c}{\textbf{LMA-8}}\\
 \hline
 \multirow{5}*{Ope} & \multirow{2}*{Student 1} & Ppl. & 31.84 $\pm$0.31 & 31.89 $\pm$0.64 & 30.91 $\pm$0.43 & 30.80 $\pm$0.39 & \textbf{30.21 $\pm$0.25} & 5.1\% $\downarrow$\\
 \cline{3-9}
 \multirow{5}*{443.4 MB} & \multirow{2}*{177.6 MB} & BLEU & 13.73 $\pm$0.19 & 13.67 $\pm$0.27 & 13.89 $\pm$0.26 & 13.98 $\pm$0.21 & \textbf{14.11 $\pm$0.12} & 2.8\% $\uparrow$\\
 \cline{3-9}
 \multirow{5}*{BLEU 14.92} & ~ & Mem. & 407.39 & 458.98 (+52) & 430.77 (+23) & 719.73 (+312) & 487.23 (+80) & 74\%~~$\downarrow$\\
 \hhline{~|========}
 \multirow{5}*{Ppl. 29.71} & \multirow{2}*{Student 2} & Ppl. & 44.51 $\pm$0.52 & 44.23 $\pm$0.56 & 43.44 $\pm$0.39 & 42.97 $\pm$0.62 & \textbf{41.21 $\pm$0.35} & 7.4\% $\downarrow$\\
 \cline{3-9}
 \multirow{5}*{Mem.1014.8} & \multirow{2}*{87.2 MB} & BLEU & 10.46 $\pm$0.18 & 10.51 $\pm$0.24 & 10.78 $\pm$0.23 & 10.87 $\pm$0.30 & \textbf{10.94 $\pm$0.18} & 4.6\% $\uparrow$\\
 \cline{3-9}
 ~ & ~ & Mem. & 282.05 & 335.34 (+53) & 305.43 (+23) & 596.10 (+314) & 363.60 (+82) & 74\%~~$\downarrow$\\
 \hhline{~|========}
 ~ & \multirow{2}*{Student 3} & Ppl. & 71.69 $\pm$0.51 & 72.56 $\pm$1.03 & 70.45 $\pm$0.69 & 70.31 $\pm$0.61 & \textbf{67.62 $\pm$0.31} & 5.7\% $\downarrow$\\
 \cline{3-9}
 ~ & \multirow{2}*{43.3 MB} & BLEU & ~~6.12 $\pm$0.12 & ~~6.06 $\pm$0.15 & ~~6.26 $\pm$0.25 & ~~6.40 $\pm$0.29 & \textbf{~~6.64 $\pm$0.04} & 8.5\% $\uparrow$\\
  \cline{3-9}
 ~ & ~ & Mem. & 220.49 & 274.63 (+54) & 243.87 (+23) & 535.39 (+315) & 302.89 (+82) & 74\%~~$\downarrow$\\
 \hhline{=========}
 \multirow{2}*{WMT13} & \multirow{2}*{Student 1} & Ppl. & ~~6.44 $\pm$0.02 & ~~6.47 $\pm$0.03 & ~~6.34 $\pm$0.03 & \multirow{3}*{OOM} & \textbf{~~6.29 $\pm$0.04} & 2.3\% $\downarrow$\\
 \hhline{~~----~--}
 \multirow{2}*{443.4 MB} & \multirow{2}*{177.6 MB} & BLEU & 26.89 $\pm$0.05 & 26.81 $\pm$0.06 & 26.98 $\pm$0.08 & ~ & \textbf{27.12 $\pm$0.07} & 0.9\% $\uparrow$\\
 \hhline{~~----~--}
 \multirow{2}*{BLEU 28.56} & ~ & Mem. & 419.40 & 470.99 (+52) & 442.78 (+23) & ~ & 499.24 (+81) & N/A\\
 \hhline{~|========}
 \multirow{2}*{Ppl. 5.31} & \multirow{2}*{Student 2} & Ppl. & 12.61 $\pm$0.05 & 12.72 $\pm$0.04 & 12.51 $\pm$0.03 & 12.35 $\pm$0.06 & \textbf{12.25 $\pm$0.05} & 2.9\% $\downarrow$\\
 \hhline{~~-------}
 \multirow{2}*{Mem. 1040.8} & \multirow{2}*{43.3 MB} & BLEU & 20.39 $\pm$0.09 & 19.96 $\pm$0.07 & 20.82 $\pm$0.08 & 21.02 $\pm$0.10 & \textbf{21.19 $\pm$0.08} & 3.9\% $\uparrow$\\
 \hhline{~~-------}
 ~ & ~ & Mem. & 230.83 & 284.97 (+54) & 254.21 (+23) & 545.73 (+315) & 313.23 (+82)& 74\%~~$\downarrow$\\
\bottomrule
\end{tabular}
\label{tab:MT-res}
\end{table*}

\subsection{Machine Translation}

\subsubsection{Setting}
To further evaluate the effectiveness of LMA, we also conduct experiments on machine translation using the OpenNMT integration test dataset (Ope) consisting of 200K train sentences and 10K test sentences and WMT13 \cite{koehn2005europarl} dataset for a German-English translation task. The translational models we employed are based on the seq2seq models from OpenNMT \footnote{\url{https://github.com/OpenNMT/OpenNMT-py}}, where the encoder and decoder are both Transformers \cite{vaswani2017attention} instead of LSTM used in \cite{Polino2018Model}. LSTM is not selected because its activations are usually Sigmoid and Tanh, both of which are saturated and much different from PLA. Besides one teacher model for each data, we also employ three student models with different sizes on Ope, and two student models on WMT13.
We use the perplexity (\textit{Ppl.}, lower is better) and the BLEU score (\textit{BLEU}), computed by the Moses project (mos), as two evaluation metrics.

\subsubsection{Result}
Table~\ref{tab:MT-res} shows the results on machine translation. From this table, we can find that our method outperforms all the baseline activations. Specifically, the BLEU scores of LMA increase by 3\% to 8\% over ReLU on Ope and 1\% to 4\% on larger WMT13. Moreover, we can observe the similar advantages of LMA in terms of the multi-segment effectiveness and memory cost comparison as in image classification tasks.
It is worth to note that using APLU may cause memory overflow due to its huge cost (Out of Memory, OOM), as shown by APLU-equipped Student-1 on WMT13.

\begin{figure}
    \centering
    \includegraphics[width=0.8\linewidth]{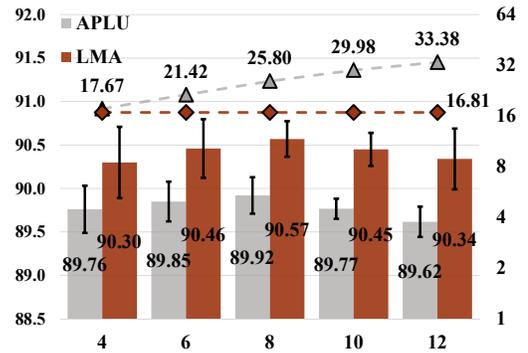}
    \caption{Segment Study for APLU and LMA on CIFAR-10. The bars and left axis show accuracy (\%) while the lines and right axis show memory cost (MB). \textbf{The cost of APLU grows linearly but that of LMA remains much lower.}}
    \label{fig:bins}
\end{figure}
\subsection{Additional Experiment}

\begin{table*}[]
    \centering
    \small \setlength\tabcolsep{4.6pt}
    \caption{Joint Use Results with Quantized Distillation on CIFAR-10. The Teacher model employed is the same as the one in the above experiments on CIFAR-10. \textbf{It shows that LMA works well with Quantization and Distillation, at the same time.}}
    \begin{tabular}{c|c c | c c | c c}
    \toprule
    \multirow{2}*{\textbf{Method}} & \multicolumn{2}{c|}{\textbf{Student 1}} & \multicolumn{2}{c|}{\textbf{Student 2}} & \multicolumn{2}{c}{\textbf{Student 3}} \\
    ~ & \textbf{ReLU} & \textbf{LMA-8} & \textbf{ReLU} & \textbf{LMA-8} & \textbf{ReLU} & \textbf{LMA-8}\\
    \hline
    4 bits
    & 85.74 $\pm$0.15 & \textbf{86.31 $\pm$0.41} & 77.04 $\pm$0.51 & \textbf{79.48 $\pm$0.79} & 65.33 $\pm$0.17 & \textbf{68.85 $\pm$0.99} \\
    \hline
    8 bits
    & 87.02 $\pm$0.23 & \textbf{88.56 $\pm$0.52} & 80.53 $\pm$0.75 & \textbf{83.37 $\pm$0.51} & 70.23 $\pm$0.98 & \textbf{74.47 $\pm$0.74} \\
    \bottomrule
    \end{tabular}
    \label{tab:joint}
\end{table*}

\subsubsection{Segment Study}
To verify if the expressiveness ability can be enhanced via increasing the segment number, we conduct additional experiments on CIFAR-10 to study the effect of segment number $k$ in LMA. As shown in Fig.~\ref{fig:bins}, with the segment number increasing from 4 to 8, both APLU and LMA yield soaring performance. Despite a slight decline beyond 10, LMA is still much better than ReLU.
Besides, the memory cost of APLU grows linearly with the segment number while that of LMA remains stable and much lower.

\subsubsection{Joint Use}
To show the effectiveness of the jointly using our method with other compression techniques, we conduct further experiment to combine Quantized Distillation \cite{Polino2018Model} with our method on CIFAR-10. 
From Table~\ref{tab:joint}, we can find that the accuracy of LMA-equipped model is much higher than that of ReLU-equipped one, also by about 2\% to 6\%, with all different settings of the number of bits in the quantized model.


Overall, all experiments above have implied that the multi-segment activation, including APLU and LMA, can achieve better performance than the two-segment ones, and the improvement brought by multi-segment design becomes increasingly apparent against reducing model size. Therefore, it is quite useful to leverage the segment number of PLA to improve the performance of the compact model in model compression.
Furthermore, LMA outperforms APLU mostly and maintain more efficient memory usage simultaneously even in the only one case LMA not beating APLU. 
It indicates that the high efficiency of LMA makes it quite suitable in resources-exhausted environments. More than this, LMA can also be used conveniently and effectively together with the other techniques. To conclude, LMA can play the most critical role in model compression due to its highly competitive effectiveness, efficiency and compatibility.

\section{Conclusion and Outlook}

In model compression, especially knowledge distillation, to fill the expressiveness ability gap between the compact NN and complex NN, we propose a novel highly efficient Light Multi-segment Activation (LMA) in this paper, which empowers the compact NN to yield comparable performance with the complex one.
Specifically, to produce more segments but preserving low resource cost, LMA uses statistical information of the batch input to determine multiple segment cut points. Then, it transforms the inputs linearly over different segments.
Experimental results on the real-world tasks with multi-size NN have demonstrated the effectiveness and efficiency of LMA. Besides, LMA is well compatible with the other techniques like quantization, also helping the performance of other approaches improved.

To the best of our knowledge, it is the first work that leverages multi-segment piecewise linear activation for model compression, which provides a good insight on designing efficient and powerful compact models. In the future, on the one hand, we will further reduce the time and space costs of LMA computing from the bottom as much as possible, by hardware-level or specialized computation. On the other hand, improving the capacity of activation is also a novel and significant direction to simplify complex architectures and apply Neural Networks more efficiently.

\bibliographystyle{aaai}
{\footnotesize
\bibliography{2073-refs}}
\begin{appendices}

\begin{table*}
    \centering
    \caption{Model specifications on CIFAR-10.}
    \vspace{-5pt}
    \begin{tabular}{c c c}
    \toprule
    ~ & Architecture & Parameter Number \\
    \midrule
    Teacher Model & 76$c^2$-mp-dp-126$c^2$-mp-dp-148$c^4$-mp-dp-1200fc-dp-1200fc & 5.34 M\\
    Student Model 1& 75c-mp-dp-50$c^2$-mp-dp-25c-mp-dp-500fc-dp & 1.01 M\\
    Student Model 2& 50c-mp-dp-25$c^2$-mp-dp-10c-mp-dp-400fc-dp & 0.32 M\\
    Student Model 3& 25c-mp-dp-10$c^2$-mp-dp-5c-mp-dp-300fc-dp & 0.11 M\\
    \bottomrule
    \end{tabular}
    \label{tab:cifar10-model}
\end{table*}

\section{Reproducibility Details}
We anonymously released the source code at: \url{https://github.com/LMA-NeurIPS19/LMA}, where all of the experimental codes and method implementations exist, and it is mainly built from the codebase\footnote{\url{https://github.com/antspy/quantized_distillation}} of  \cite{Polino2018Model}. Furthermore, we use this supplementary material to provide some important details about some specific settings, and some intuitive results in an example figure.
On the one hand, due to the state-of-the-art complex models applied in general scenarios have good enough expressiveness ability, the gain from equipping LMA is sometimes incremental for them. On the other hand, LMA is designed particularly for the compact model in model compression. So the experiments in this paper mainly focus on evaluating the performance of LMA-based compact models in compression scenario. 

Generally, in our experiments, the implementation is based on Pytorch, and all the experiments are running on NVIDIA Tesla P40, whose memory is 24 GB. Besides, the inference memory cost is recorded by the Pytorch API: \textit{torch.cuda.max\_memory\_allocated()}, when feeding the streaming test samples one-by-one, i.e. the test batch size is set to one.

\subsection{Baseline Details}

We compare the performance of LMA with several widely-used and state-of-the-art activations:

\begin{compactitem}
    \item Rectified Linear Unit (ReLU) \cite{nair2010rectified}, without hyperparameter settings.
    \item Parametric ReLU (PReLU) \cite{he2016deep} with initial slope $\alpha=0.25$, as the author suggested.
    \item Swish \cite{ramachandran2017swish}, whose equation is $h(x_i) = x \cdot \sigma (\beta x) ~$, where $\sigma( \cdot )$ is the Sigmoid function and $\beta$ is either a constant or a trainable parameter. It is a state-of-the-art activation but not PLA essentially, but we can treat it as an approximation to two-segment ReLU. We use the version outperforming in \cite{ramachandran2017swish}, that the parameter $\beta$ is trainable and initialized to one.
    \item Adaptive Piecewise Linear Units (APLU) \cite{agostinelli2014learning}. The segment number of APLU is set to the same as LMA. The initialization of APLU is according to the author's code \cite{Agostinelli2015}, that the slopes are initialized uniformly and the cut points normally. Besides, it is worth mentioning that the $k$ set in the equation of APLU (see Eqn.~1 in the paper) is not consistent with its segment number we claimed, that it denotes the number of cut points. Therefore, an APLU with $k$ boundaries has $k+2$ segments totally, with $k+1$ dynamic segments and one extra from the ReLU added. For a fair comparison, when we mentioned a \textit{APLU-K}, exactly we set $K-2$ cut points in it.
\end{compactitem}

We do not take Maxout as the baseline because its parameter size is obviously much huger than the others. Besides, in LMA, the moving average factor, for updating segment cut points $b$ in the inference phase, is set to 0.99. Except for the segment study experiments, the segment number is usually set to 8, which is the same as that in APLU.

\subsection{Dataset Details}
For image classification, CIFAR-10 and CIFAR-100 \cite{krizhevsky2009learning} are used. Both CIFAR-10 and CIFAR-100 are the well-known image classification benchmark datasets, which contain 50K training set and 10K testing set, and where images contain $32 \times 32 \times3$ pixels. The difference between them is that there are 10-classes labels in CIFAR-10 while 100-classes in CIFAR-100.

For machine translation, we evaluate our method on the OpenNMT integration test dataset (Ope) consists of 200K train sentences and 10K test sentences\footnote{Obtained like \url{https://github.com/antspy/quantized_distillation/blob/master/datasets/translation_datasets.py\#L211}}, and well-known WMT13 \cite{koehn2005europarl} dataset for a German-English translation task. The WMT13 used contains 1.7M training sentences and 190K testing sentences.

\subsection{Model Details}

On CIFAR-10, the model specifications are listed in Table~\ref{tab:cifar10-model}, where $c^a$ denotes $a$ convolutional layers, \textit{mp} denotes max pooling layer, \textit{dp} denotes dropout layer and \textit{fc} denotes fully-connected layer.

On CIFAR-100, the parameter settings for the structure of Wide Residual Networks\footnote{Implementation refers to \url{https://github.com/meliketoy/wide-resnet.pytorch}} are listed in Table~\ref{tab:cifar100-model}. The detailed architecture of Wide Residual Networks (WRN) refers to \cite{zagoruyko2016wide}, where the meaning of the listed parameters is also shown. Additionally, we employ relatively shallow WRN-16 as the teacher model, because with the depth increases more, the performance of WRN is not improved obviously anymore (the Error metric improves from 21.59\% only to 20.75\%, with depth from 16 to 22 reported in \cite{zagoruyko2016wide}).

\begin{table}
    \centering
    \small
\setlength\tabcolsep{3.6pt}
    \caption{Model specifications on CIFAR-100.}
    \vspace{-5pt}
    \begin{tabular}{cccc}
    \toprule
    Model & Widen Factor & Depth & Parameter Number \\
    \hline
    Teacher Model & 10 & 16 & 17.2 M\\
    Student Model 1 & 6 & 10 & 1.22 M\\
    Student Model 2 & 4 & 10 & 0.32 M\\
    \bottomrule
    \end{tabular}
    \label{tab:cifar100-model}
\end{table}

For machine translational models, we employ multi-layer transformers \cite{vaswani2017attention} as the encoder and decoder in seq2seq framework, to evaluate the effectiveness of our method.
The implementation is based on OpenNMT-py\footnote{\url{https://github.com/OpenNMT/OpenNMT-py}} and Pytorch, and the model specifications are listed in Table~\ref{tab:MT-model}. On Ope, all of the four models are running, while the teacher model, the first and the last student model are selected to evaluate on WMT13, for the space limitation and huge time cost. Additionally, the BLEU score is computed by \textit{multi-bleu.perl} code from the moses project (mos)\footnote{\url{http://www.statmt.org/moses/?n=moses.baseline}}.

\begin{table*}
    \centering
    \caption{Model specifications for machine translation.}
    \vspace{-5pt}
    \begin{tabular}{cccccc}
    \toprule
    Model & Embedding Size & Hidden Units & Encoder Layers & Decoder Layers & Parameter Number \\
    \hline
    Teacher Model & 512 & 512 & 6 & 6 & 116 M\\
    Student Model 1 & 256 & 256 & 3 & 3 & 47 M\\
    Student Model 2 & 128 & 128 & 3 & 3 & 23 M\\
    Student Model 3 & 64 & 64 & 3 & 3 & 11 M\\
    \bottomrule
    \end{tabular}
    \label{tab:MT-model}
\end{table*}

\subsection{Hyper-parameter Setting}

As the distributions of the used datasets in experiments are different from each other, we use different hyper-parameters on different datasets. However, the settings are only different between the datasets, without varying on the models, which means the parameter setting is strictly the same in all of the models when on one specific dataset. Besides, all the hyper-parameters are basically set to be the same as the \cite{Polino2018Model}'s settings, without deliberate adjustments.

We list the hyper-parameters on CIFAR-10 in Table~\ref{tab:cifar10-hp}.
The learning rate decay strategy is according to the implementation in \cite{Polino2018Model}, where the adjustment of learning rate depends on the changing trend of validation accuracy.
If the validation accuracy does not increase anymore, after waiting for a fixed epoch, the learning rate is halved. On CIFAR-100, the hyper-parameters are listed in Table~\ref{tab:cifar100-hp}, and its learning rate adjustment is the same as the setting in \cite{zagoruyko2016wide}.

The hyper-parameters for machine translation are listed in Table~\ref{tab:MT-hp}. The model and hyper-parameters setting are the same on both Ope and WMT-13, which are mainly set to official default settings recommended by OpenNMT. More details can be found in our codes, the \textit{standard\_options.py} file in \textit{onmt} directory specifically.

\begin{table*}
    \centering
    \caption{Hyper-parameters on CIFAR-10.}
    \vspace{-5pt}
    \begin{tabular}{C{0.15\textwidth}|C{0.35\textwidth}|C{0.3\textwidth}}
    \toprule
    \multicolumn{2}{c|}{Batch Size} & 64 \\
    \hline
    \multicolumn{2}{c|}{Maximum Epoch} & 200 \\
    \hline
    \multicolumn{2}{c|}{Batch Normalization} & True \\
    \hline
    \multicolumn{2}{c|}{Weight Decay} & 2.2e-4 \\
    \hline
    \multirow{5}*{Learning Rate} & Initial LR & 0.01 \\
    ~ & Decay Factor & 0.5 \\
    ~ & Epoch to Wait Before Decaying & 10 \\
    ~ & Epoch to Wait After Decaying & 8 \\
    ~ & Maximum Decay Times & 11 \\
    \hline
    \multirow{2}*{Optimizer} & Method & Stochastic Gradient Descent \\
    ~ & Momentum & 0.9 \\
    \hline
    \multirow{2}*{Distillation} & Distillation Loss Weight & 0.7 \\
    ~ & Soften Temperature & 2\\
    \bottomrule
    \end{tabular}
    \label{tab:cifar10-hp}
\end{table*}

\begin{table*}
    \centering
    \caption{Hyper-parameters on CIFAR-100.}
    \vspace{-5pt}
    \begin{tabular}{C{0.2\textwidth}|C{0.3\textwidth}|C{0.3\textwidth}}
    \toprule
    \multicolumn{2}{c|}{Batch Size} & 128 \\
    \hline
    \multicolumn{2}{c|}{Epoch} & 200 \\
    \hline
    \multicolumn{2}{c|}{Batch Normalization} & True \\
    \hline
    \multicolumn{2}{c|}{Weight Decay} & 5e-4 \\
    \hline
    \multicolumn{2}{c|}{Dropout Rate} & 0.3 \\
    \hline
    \multirow{4}*{Learning Rate} & LR in Epoch 1-60 & 0.1 \\
    ~ & LR in Epoch 61-120 & 0.02 \\
    ~ & LR in Epoch 121-160  & 4e-3 \\
    ~ & LR in Epoch 161-200  & 8e-4 \\
    \hline
    \multirow{2}*{Optimizer} & Method & Stochastic Gradient Descent \\
    ~ & Momentum & 0.9 \\
    \hline
    \multirow{2}*{Distillation} & Distillation Loss Weight & 0.7 \\
    ~ & Soften Temperature & 2\\
    \bottomrule
    \end{tabular}
    \label{tab:cifar100-hp}
\end{table*}

\begin{table*}
    \centering
    \caption{Hyper-parameters for machine translation.}
    \vspace{-5pt}
    \begin{tabular}{C{0.25\textwidth}|C{0.25\textwidth}|C{0.3\textwidth}}
    \toprule
    \multicolumn{2}{c|}{Epoch} & 15 \\
    \hline
    \multicolumn{2}{c|}{Batch Normalization} & True \\
    \hline
    \multicolumn{2}{c|}{Dropout Rate} & 0.1 \\
    \hline
    Attention Mechanism & Head Count & 8 \\
    \hline
    \multirow{2}*{Batch Setting} & Batch Type ~ & Tokens \\
    ~ & Batch Size & 3192 \\
    \hline
    \multirow{4}*{Learning Rate} & Initial LR & 2.0 \\
    ~ & Decay Factor & 0.5 \\
    ~ & Start Decay at Epoch & 8 \\
    ~ & Decay Method & noam \\
    \hline
    \multirow{2}*{Optimizer} & Method & Adam \\
    ~ & beta & 0.998 \\
    \hline
    \multirow{2}*{Distillation} & Distillation Loss Weight & 0.7 \\
    ~ & Soften Temperature & 1\\
    \hline
    Beam Computing & Beam Size & 5 \\
    \bottomrule
    \end{tabular}
    \label{tab:MT-hp}
\end{table*}

\begin{figure}
    \centering
    \includegraphics[width=0.45\textwidth]{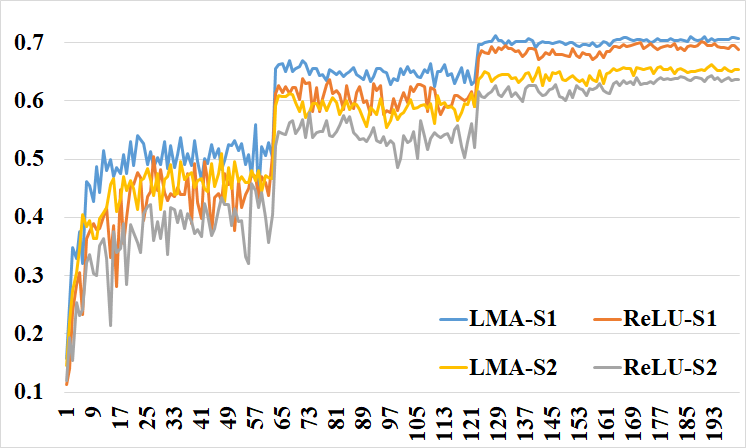}
    \caption{Testing Accuracy-Epoch Curves on CIFAR-100. The vertical axis is accuracy (\%) and the horizontal one is epoch. ``S1'' and ``S2'' means Student-1 and Student-2 respectively.}
    \label{fig:cifar100}
\end{figure}

\subsection{Accuracy-Epoch Curves on CIFAR-100}
To show the effectiveness of LMA more intuitively, we provide one more example figure here, shows some Testing Accuracy Curves of the student models based on ReLU and LMA-8 respectively, when training on CIFAR-100. From the curves in Fig.~\ref{fig:cifar100}, it is easily to find the high effectiveness of LMA from its much improvement from ReLU.

\end{appendices}
\end{document}